# A Mosquito is Worth 16x16 Larvae: Evaluation of Deep Learning Architectures for Mosquito Larvae Classification

Aswin Surya, David B. Peral, Austin VanLoon, Akhila Rajesh

**Abstract** Mosquito-borne diseases (MBDs), such as dengue virus, chikungunya virus, and West Nile virus, cause over one million deaths globally every year. Because many such diseases are spread by the *Aedes* and *Culex* mosquitoes, tracking these larvae becomes critical in mitigating the spread of MBDs. Even as citizen science grows and obtains larger mosquito image datasets, the manual annotation of mosquito images becomes ever more time-consuming and inefficient. Previous research has used computer vision to identify mosquito species, and the Convolutional Neural Network (CNN) has become the de-facto for image classification. However, these models typically require substantial computational resources. This research introduces the application of the Vision Transformer (ViT) in a comparative study to improve image classification on *Aedes* and *Culex* larvae. Two ViT models, ViT-Base and CvT-13, and two CNN models, ResNet-18 and ConvNeXT, were trained on mosquito larvae image data and compared to determine the most effective model to distinguish mosquito larvae as *Aedes* or *Culex*. Testing revealed that ConvNeXT obtained the greatest values across all classification metrics, demonstrating its viability for mosquito larvae classification. Based on these results, future research includes creating a model specifically designed for mosquito larvae classification by combining elements of CNN and transformer architecture.

## I. INTRODUCTION

Among Earth's most deadly species, mosquitoes are among the deadliest. In fact, mosquito-borne diseases (MBDs) are responsible for at least 725,000 deaths annually [16]. While MBDs have challenged humans for generations, factors like urbanization, climate change, and population growth have only exacerbated the issue [20].

MBDs are especially dangerous because mosquitoes transmit viruses easily. The three most common genera of mosquitos are the *Aedes*, *Culex*, and *Anopheles* mosquitoes. *Aedes* and *Culex* mosquitoes are especially deadly because they can breed anywhere, not just in natural environments. For instance, the female *Aedes aegypti* mosquito can lay eggs in any moist and warm environment. In Brazil, the *Aedes aegypti* species alone started a Zika epidemic and caused 2,500 cases of microcephaly, a condition where a baby's brain has not developed properly [12]. Additionally, *Culex* mosquitoes rapidly spread the West Nile virus, the leading cause of MBD in the United States [4].

The spread of MBDs can be prevented by classifying mosquito larvae, which are distinguishable by their siphon. Larvae classification allows health officials to track mosquito populations in an area, learn which species thrive in certain environments, and forecast the presence of invasive species to prevent outbreaks, as only certain mosquito species transmit certain viruses [11]. This preventative approach is important because many MBDs like dengue virus and West Nile Virus, which has become endemic in the U.S., have no vaccine or treatment [3]–[17].

Recent research has focused on the use of artificial intelligence (AI) as an alternative to manual classification. For image classification, convolutional neural networks (CNNs) and vision transformers (ViTs) are the most common. CNNs perform convolution to locally extract features from an image and produce a feature map, from which the network can classify an image. Many past works have previously applied CNNs to identify mosquito-specific tasks. Reference [9] achieved an accuracy of 89.50% when identifying unknown mosquito species and 88.72% when

The NASA Earth Science Education Collaborative leads Earth Explorers through an award to the Institute for Global Environmental Strategies, Arlington, VA (NASA Award NNX6AE28A). The SEES High School Summer Intern Program is led by the Texas Space Grant Consortium at the University of Texas at Austin (NASA Award NNX16AB89A).

A. Surya is with Bellarmine College Preparatory, San Jose, CA 95126, USA (e-mail: aswinsurya@gmail.com).

D. B. Peral is with La Cañada High School, Flintridge, CA 91011, USA (e-mail: davidbperal@gmail.com).

A. VanLoon is with Middleton High School, Tampa, FL 33610, USA (e-mail: aavanloon@gmail.com).

A. Rajesh is with Westwood High School, Austin, TX 78750, USA (e-mail: akhila.rrajesh@gmail.com).

identifying known species with CNNs. Reference [8] compared CNNs with the You-Only-Look-Once (YOLO) algorithm to predict mosquito habitats, and the YOLOv4 CNN worked best. The study concluded that CNNs were the most efficient and cost-effective approach to predict large-scale mosquito habitats but were unable to identify small-scale habitats such as footprints, tires, or puddles.

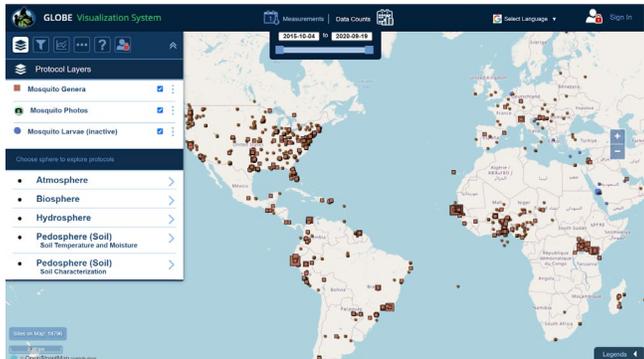
Fig. 1. The GLOBE Mosquito Habitat Mapper database, used to access the larvae data.

While CNNs were the longtime state-of-the-art machine learning for image classification, [6] proposed a novel architecture that outperformed CNNs: the ViT.

Rather than convolutions, the ViT uses self-attention to integrate all features of the data. Unlike CNNs, ViTs have not been extensively applied to mosquito-related tasks. Thus far, [18] reached an accuracy of 90.03% in using ViTs to predict malaria using thin blood smear microscopic images.

This work seeks to compare the CNN to the ViT for mosquito larvae classification to proactively prevent the spread of MBDs. This study compares four machine learning models to classify larvae as *Aedes*, *Culex*, or neither, a distinction from many previous works.

## II. METHODS

### A. Mosquito Larvae Dataset

All mosquito larvae data were obtained using the GLOBE Mosquito Habitat Mapper database (depicted in Fig. 1 above), an app-based citizen science initiative to collect nationwide data on mosquito habitats and identification. Data were collected from May 31, 2017, to July 7, 2022, from North America, Latin America, and Africa.

### B. Data Preparation

This work required a multistep data preparation process to maintain the quality, accuracy, and consistency of the data. The data was downloaded using a spreadsheet builder and then converted into comma-separated values (CSV) files. Each image had to be carefully classified as *Aedes*, *Culex*, or neither, based on the length and shape of the siphon, color of the larvae, and amount of hair. The main feature that distinguishes larvae is their siphon, the organ used to breathe. *Aedes* larvae have a shorter and wider siphon, while *Culex* larvae have a longer and thinner siphon. After classification, the additional commas in the "location" column needed to be removed due to the data loader assuming that those were extraneous blank columns.

Pre-processing also included removing several image links that were no longer supported by the server. When loading the image data, a separate script was used to identify and remove URLs with erroneous HTTP status codes. Finally, null values were also reviewed and validated.

After cleaning the data, the dataset was uploaded to the HuggingFace Hub to be easily imported into code. The train data was the mosquito larvae data from Africa, consisting of 7107 rows, whereas the test data was from North America and Latin America, consisting of 3439 rows. The image links were cast as PIL image data and converted to pixel values. Prior to training each model, feature extraction was performed on the data using the Feature Extractor class in the HuggingFace transformers library. Finally, the classification head of each model was altered to three to account for the three individual labels.

### C. Convolutional Neural Network

A Convolutional Neural Network (CNN) is a deep learning (DL) architecture widely used for image classification that extracts local features and learns directly from them using kernel convolutions. The kernel convolution is the process of taking a small matrix of numbers, called a filter, and passing it over the image in a specified stride and padding. This process repeats, constantly sliding until the entire image has been covered. The result of this process is a feature map, from which the network can comprehend features and make a classification. The formula to calculate the dimensions of the output matrix, $n_{out}$, where $p$ is the padding, $f$ is the filter dimension, $s$ is the stride, and $n_{in}$ is the input image is:

$$n_{out} = \frac{n_{in} + 2p - f}{s} + 1$$

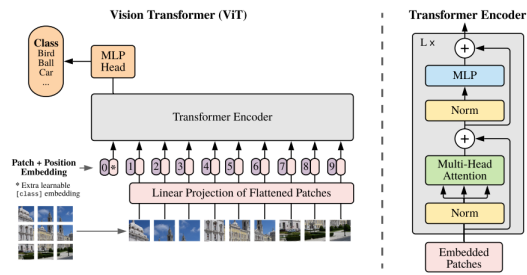
Fig. 2. A side-by-side comparison of the Vision Transformer and the Transformer Encoder. The attention-based Transformer Encoder is utilized in the ViT, but it also includes a classification token and multi-layer perceptron to perform class prediction. Figure acquired from [2].

$$\text{MultiHead}(\mathbf{Q}, \mathbf{K}, \mathbf{V}) = \text{Concat}(\text{head}_1, \ldots, \text{head}_h)\mathbf{W}^O$$
$$\text{where head}_i = \text{Attention}(\mathbf{Q}\mathbf{W}_i^Q, \mathbf{K}\mathbf{W}_i^K, \mathbf{V}\mathbf{W}_i^V)$$

Fig. 3. Multi-head attention using query, key, and values. Each independent head is concatenated and transformed using a square weight matrix,

allowing the model to perform self-attention computation in parallel. Figure acquired from [1].

*D. Vision Transformer*

Transformers were first introduced as a novel approach to Natural Language Processing (NLP) using a sequential, purely attention-based mechanism [21]. Later, the transformer was applied to computer vision, creating the vision transformer (ViT) [6]. Inspiring the title of this paper, the ViT sought to utilize nothing but attention on sequential patches of images. It splits an input image into patches and uses positional embeddings to capture the relative location of the patch. The ViT uses a mechanism known as self-attention instead of convolutions to integrate all features of the data across the model, from the lowest to highest layers, enabling the network to learn the local and global features of the image (see Fig. 2).

The ViT architecture also employs multi-head attention, running through the attention mechanism several times (see Fig. 3). The query, key, and value matrices are mapped into lower-dimensional spaces, and attention is continually computed, with each output referred to as a head. The independence of each head allows for parallelization of self-attention computation, facilitating the model to attend to different parts of the sequence differently and capture unique positional information.

*E. ResNet-18*

One of the most-employed CNN models is ResNet, which adopted residual learning to stacked layers. Deep CNNs face the vanishing gradient problem, where gradients calculated from the loss function shrink to zero because of repeated applications of the chain rule, hindering training in early layers. This saturates the network's accuracy and increases the network's training error, effectively reducing its generalization capability. ResNet adds skip connections to allow the gradients to flow from the final layers to the initial filters. The skip connections with residual mapping also allowed ResNets to more easily optimize and gain accuracy from increased depth, removing the previous bottleneck. ResNets were shown to perform better than comparative CNNs such as VGG, GoogLeNet, and Inception networks [10].

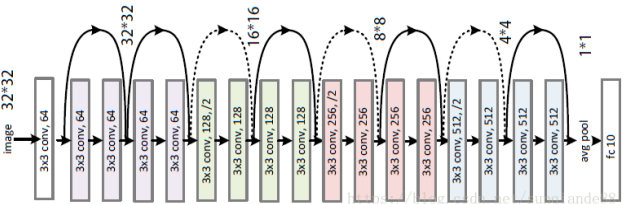

Fig. 4. The ResNet-18 with 18 parameter layers with inserted shortcut connections for performing identity mapping. Table 1 has more details for other architectures. Figure acquired from [7].

ResNet-18 is the CNN architecture of choice in this work as it is a widely-used CNN benchmark in image classification. The ResNet-18 used was pre-trained on ImageNet-1k and fine-tuned on the mosquito larvae dataset.

| layer name | output size | 18-layer | 34-layer | 50-layer | 101-layer | 152-layer |
|---|---|---|---|---|---|---|
| conv1 | 112×112 | colspan 7×7, 64, stride 2 | | | | |
| | | colspan 3×3 max pool, stride 2 | | | | |
| conv2_x | 56×56 | $\begin{bmatrix}3\times3, 64\\3\times3, 64\end{bmatrix}\times2$ | $\begin{bmatrix}3\times3, 64\\3\times3, 64\end{bmatrix}\times3$ | $\begin{bmatrix}1\times1, 64\\3\times3, 64\\1\times1, 256\end{bmatrix}\times3$ | $\begin{bmatrix}1\times1, 64\\3\times3, 64\\1\times1, 256\end{bmatrix}\times3$ | $\begin{bmatrix}1\times1, 64\\3\times3, 64\\1\times1, 256\end{bmatrix}\times3$ |
| conv3_x | 28×28 | $\begin{bmatrix}3\times3, 128\\3\times3, 128\end{bmatrix}\times2$ | $\begin{bmatrix}3\times3, 128\\3\times3, 128\end{bmatrix}\times4$ | $\begin{bmatrix}1\times1, 128\\3\times3, 128\\1\times1, 512\end{bmatrix}\times4$ | $\begin{bmatrix}1\times1, 128\\3\times3, 128\\1\times1, 512\end{bmatrix}\times4$ | $\begin{bmatrix}1\times1, 128\\3\times3, 128\\1\times1, 512\end{bmatrix}\times8$ |
| conv4_x | 14×14 | $\begin{bmatrix}3\times3, 256\\3\times3, 256\end{bmatrix}\times2$ | $\begin{bmatrix}3\times3, 256\\3\times3, 256\end{bmatrix}\times6$ | $\begin{bmatrix}1\times1, 256\\3\times3, 256\\1\times1, 1024\end{bmatrix}\times6$ | $\begin{bmatrix}1\times1, 256\\3\times3, 256\\1\times1, 1024\end{bmatrix}\times23$ | $\begin{bmatrix}1\times1, 256\\3\times3, 256\\1\times1, 1024\end{bmatrix}\times36$ |
| conv5_x | 7×7 | $\begin{bmatrix}3\times3, 512\\3\times3, 512\end{bmatrix}\times2$ | $\begin{bmatrix}3\times3, 512\\3\times3, 512\end{bmatrix}\times3$ | $\begin{bmatrix}1\times1, 512\\3\times3, 512\\1\times1, 2048\end{bmatrix}\times3$ | $\begin{bmatrix}1\times1, 512\\3\times3, 512\\1\times1, 2048\end{bmatrix}\times3$ | $\begin{bmatrix}1\times1, 512\\3\times3, 512\\1\times1, 2048\end{bmatrix}\times3$ |
| | 1×1 | colspan average pool, 1000-d fc, softmax | | | | |
| FLOPs | | $1.8\times10^9$ | $3.6\times10^9$ | $3.8\times10^9$ | $7.6\times10^9$ | $11.3\times10^9$ |

Table I. ResNet architectures for ImageNet. Table acquired from [10].

*F. ConvNeXT*

After their success, the application of the ViT's design in CNNs was limited, leading to the ConvNeXT.

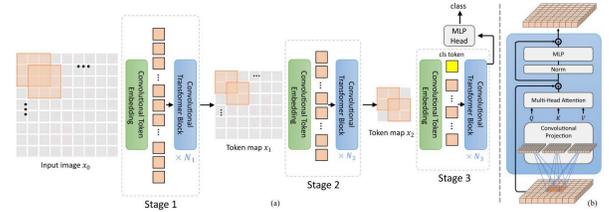

Fig. 5. CvT architecture pipeline. a) Overall architecture, demonstrating hierarchical structure with the Convolutional Token Embedding layer. b) Expanded view of the Convolutional Transformer Block, containing convolutional projection. Figure acquired from [22].

Beginning with ResNet-50, [14] gradually modernized the architecture to the design of a transformer by training the ResNet using vision transformer techniques like the AdamW Optimizer, extending epochs to 300 instead of the usual 90, and several regularization schemes. ConvNeXT replaced the ResNet's 4x4 downsampling stem cell with a "patchify" strategy found in Vision Transformers, with larger kernel size and non-overlapping convolution. Depthwise convolutions were also utilized at the same time, increasing the network width. The ReLu activation function was replaced with the Gaussian Error Linear Unit (GELU), which drops all residual layers from the block except the one between 1x1 Conv layers, replicating the style of the Transformer block. The equation is given by:

$$GELU(x) = 0.5x(1 + \tanh(\sqrt{\frac{2}{\pi}}(x + 0.044175x^3)))$$

The ConvNeXT model used was pre-trained on ImageNet-22k at resolution 224x224 and fine-tuned on the mosquito-larvae dataset.

*G. Convolutional Vision Transformer (CvT)*

The Convolutional Vision Transformer (CvT) introduces convolutions to the vision transformer architecture (see Fig. 5). A Convolutional Token Embedding layer applies convolutions on the tokenized patches and reshapes them, effectively reducing the number of tokens and increasing their depth, similar to conventional CNNs. Additionally,

Convolutional Transformer Blocks are applied instead of the usual position-wise linear projection in the ViT, where a depth-wise separable convolution operation is employed on the query, key, and value embeddings.

The CvT-13 model used was pre-trained on ImageNet-1k at resolution 224x224, and then fine-tuned on the mosquito larvae dataset.

*H. Evaluation Metrics*

All models were evaluated using four metrics: accuracy, precision, recall, and F1 score, all calculated from four values. True positives (TP) occur when the model correctly classified the test image as it was actually labeled. False positives (FP) occur when the model categorizes an image with another, incorrect classification. False negatives (FN) occur when the model does not label a mosquito larva image with the right label. True negatives (TN) occur when the model correctly classifies another image as another label as opposed to the class considered.

Accuracy describes the total number of images correctly classified:

$$Accuracy = \frac{TP + TN}{TP + FP + FN + TN}$$

Precision measures how often the model is correct when it classifies images:

$$Precision = \frac{TP}{TP + FP}$$

Recall calculates how many actual positives the model captures out of the total "positive" images labeled:

$$Recall = \frac{TP}{TP + FN}$$

F1 Score is the harmonic mean of precision and recall, and it applies the same level of importance to both precision and recall, indicating the overall performance of the model:

$$F1\ Score = 2 \times \frac{Precision \times Recall}{Precision + Recall}$$

## III. RESULTS AND DISCUSSION

The results are shown in Table 2 below. All four models had similar performances around 60%, with much lower values than expected. This general finding could be due to the regional difference between the training and testing data. In the Africa train data, there were 3917 rows of *Aedes*, 2405 rows of *Culex*, and 785 rows of "neither." In the Americas test data, there were 2062 rows of *Aedes*, 1253 rows of *Culex*, and only 124 rows of "neither." The percentage of "neither" rows in the Africa data was 11.05% whereas the percentage in the Americas data was 3.61%. This regional difference in the number of mosquito larvae that were not *Aedes* or *Culex* might have contributed to the lack of accuracy from training to testing.

| *Algorithm Results* | | | | |
|---|---|---|---|---|
| *Model* | *Accuracy* | *Precision* | *Recall* | *F1 score* |
| *ViT-Base* | 0.6374 | 0.6061 | 0.6374 | 0.5868 |
| *CvT-13* | 0.6400 | 0.6292 | 0.6400 | 0.6209 |
| **ConvNeXT** | **0.6563** | **0.6386** | **0.6563** | **0.6355** |
| *ResNet-18* | 0.5967 | 0.6034 | 0.5967 | 0.5756 |

Table II. Results from all four algorithms. The highest performing model across all metrics is the ConvNeXT.

Another issue that was faced during the training of the models was overfitting, when a model fits exactly to its training data so it is unable to classify on new, unseen test data. One reason for overfitting could be the size of the models. The ViT-Base model had 86 million trainable parameters, the CvT-13 had 19.98 million parameters, the ConvNeXT had 89 million trainable parameters, and the ResNet-18 had 11 million parameters. These large numbers of parameters means overfitting is likely. Additionally, overfitting could also be observed during training, as the training loss kept decreasing at a steady rate, whereas the evaluation loss decreased for a short amount of time before rising up and continually increasing.

The ConvNeXT scored the highest on all four classification metrics. This was likely due to the combination of standard transformer techniques with state-of-the-art CNN models like ResNet-50. Depthwise convolutions would also play a role in the robustness of the network by providing additional network width for the data to be more integrated than the other models.

The CvT-13 model performed the second best overall, within 1.63% of the ConvNeXT. The implementation of convolutions improved performance. Although it had significantly fewer trainable parameters, it still managed to achieve similar metric values. This could be due to the multi-head attention mechanism.

Although ViTs have been shown to outperform CNNs with large datasets such as CIFAR-100 and ImageNet, the ViT-Base model resulted in an accuracy of 63.74%. ViTs generally require a large amount of data to perform well, and since the mosquito larvae dataset only had 10,546 rows of data, it could have been insufficient to train. Additionally, all models were trained for 4 epochs, whereas [6] trained the ViT for 7 epochs. This was not done due to limited computational resources. Moreover, it seems that depthwise separable convolutions are a critical component in image classification due to their ability to increase the depth of an image.

ResNet-18 had the least values across all metrics. This could be due to its significantly lower number of parameters, 11 million, resulting in underfitting, when the model is unable to capture the features of the image due to limited parameters or training time. Additionally, depthwise convolutions are not included in the ResNet architecture. Because ResNet-18 was a simpler architecture in comparison to the other three models, this result is expected. Excluding the skip connections, the ResNet-18 is simply a basic CNN model, explaining the low accuracy.

## IV. Conclusions

Although all four models performed similarly for classifying mosquito larvae images as *Aedes*, *Culex*, or neither, the metric values were much lower than expected. ConvNeXT performed with the highest accuracy among the 4 models with an accuracy of 65.63%. This is likely because of the variation of regional data from the training and testing dataset. Another factor could be the images themselves, which are often only a part of the mosquito larvae instead of the whole body. From this result, it seems that CNNs with depthwise separable convolutions perform better in complex image classification tasks than purely attention-based models.

Future research could include the integration of mosquito species as opposed to genus. Mapping could also be performed by correlating the mosquito species to the most likely disease it spreads. The dataset could be further expanded to higher-quality images that reveal the whole larvae body as opposed to a single part. Testing with various sizes of the same model would allow for a comparison of parameters and identification of the optimal model capacity for image classification. This could result in a portable tool that can quickly and efficiently classify mosquito larvae while conducting fieldwork. A novel model could also be created with the sole purpose of mosquito larvae image classification by utilizing depthwise convolutions and aspects of the transformer architecture, creating a hybrid CNN and ViT model like the CvT-13 and the ConvNeXT. During the training process of these models, a finer, more thorough analysis could be conducted using methods such as increasing the number of epochs, raising the probability of dropout, and adding specific data preparation techniques. Through this approach, epidemics can be controlled efficiently and at a rapid pace by locating and identifying potentially dangerous mosquito larvae. There are several cases where rapid mosquito larvae identification and classification are necessary for public health control. Extrapolation of key features from vision transformers and convolutional neural networks to create a more efficient model would prove as a viable, cost-effective, and autonomous approach to controlling the spread of mosquito-borne diseases.

## V. Data And Code

Data and code to replicate the results of this experiment are available at the following public Github repo: https://github.com/thenerd31/vit-cnn-mosquito-image-classification.

## Acknowledgment

The authors would like to acknowledge the support of the 2022 Earth System Explorers Research Team, NASA STEM Enhancement in the Earth Sciences (SEES) Virtual High School Internship program. Special thanks to Russanne Low, Peder Nelson, and Cassie Soeffing and Ria Jain and Matteo Kimura for guidance when preparing this work.

GLOBE Observer data were obtained from NASA and the GLOBE Program and are freely available for use in research, publications and commercial applications. GLOBE Observer data analyzed in this project are publicly available at globe.gov/globe-data (accessed on 30 June 2022). The Python code to read, analyze, and visualize GLOBE data for this article as well as the analyzed datasets are available on github.com/IGES-Geospatial (accessed on 30 June 2022). Dashboard access to Mosquito Habitat Mapper and Land Cover data is available at geospatial.strategies.org (accessed on 30 June 2022).